\documentclass[letterpaper]{article} 
\usepackage[submission]{aaai25}  
\usepackage{times}  
\usepackage{helvet}  
\usepackage{courier}  
\usepackage[hyphens]{url}  
\usepackage{graphicx} 
\usepackage{natbib}  
\usepackage{caption} 
\usepackage{algorithm}
\usepackage{algorithmic}
\usepackage{amsmath}
\usepackage{multirow}
\usepackage{multicol}

\usepackage{newfloat}
\usepackage{listings}
\DeclareCaptionStyle{ruled}{labelfont=normalfont,labelsep=colon,strut=off} 
\floatstyle{ruled}
\newfloat{listing}{tb}{lst}{}
\floatname{listing}{Listing}
\title{AttFC: Attention Fully-Connected Layer for Large-Scale Face Recognition with One GPU}

\author{Zhuowen Zheng Yain-Whar Si}

\author {
    Zhuowen Zheng,
    Yain-Whar Si,
    Xiaochen Yuan,
    Junwei Duan,
    Ke Wang,
    Xiaofan Li,
    Xinyuan Zhang,
    Xueyuan Gong\textsuperscript{\rm *}
}

\begin{document}

\maketitle

\begin{abstract}
Nowadays, with the advancement of deep neural networks (DNNs) and the availability of large-scale datasets, the face recognition (FR) model has achieved exceptional performance. However, since the parameter magnitude of the fully connected (FC) layer directly depends on the number of identities in the dataset. If training the FR model on large-scale datasets, the size of the model parameter will be excessively huge, leading to substantial demand for computational resources, such as time and memory. This paper proposes the attention fully connected (AttFC) layer, which could significantly reduce computational resources. AttFC employs an attention loader to generate the generative class center (GCC), and dynamically store the class center with Dynamic Class Container (DCC). DCC only stores a small subset of all class centers in FC, thus its parameter count is substantially less than the FC layer. Also, training face recognition models on large-scale datasets with one GPU often encounters out-of-memory (OOM) issues. AttFC overcomes this and achieves comparable performance to state-of-the-art methods. Our code is available at https://github.com/LovE-K/AttFC.
\end{abstract}

%

\section{Introduction}
\label{sc:intro}
Face recognition (FR) is important in our daily lives and is widely used in many fields. Due to the development of Deep Neural Networks (DNNs) and the release of large-scale face datasets \cite{DBLP:conf/eccv/GuoZHHG16,DBLP:conf/iccvw/AnZGXZFWQZZF21,DBLP:conf/cvpr/ZhuHDY0CZYLD021,DBLP:conf/cvpr/NechK17,DBLP:conf/icip/CaoLZ18}, FR has achieved significant progress. Larger datasets improve the model performance, yet lead to a sharp increase in model parameters. Consequently, more computational resources are needed to train a model while utilizing large-scale datasets.
\par
\begin{figure}[!t]
\centering
\includegraphics[width=0.99\columnwidth]{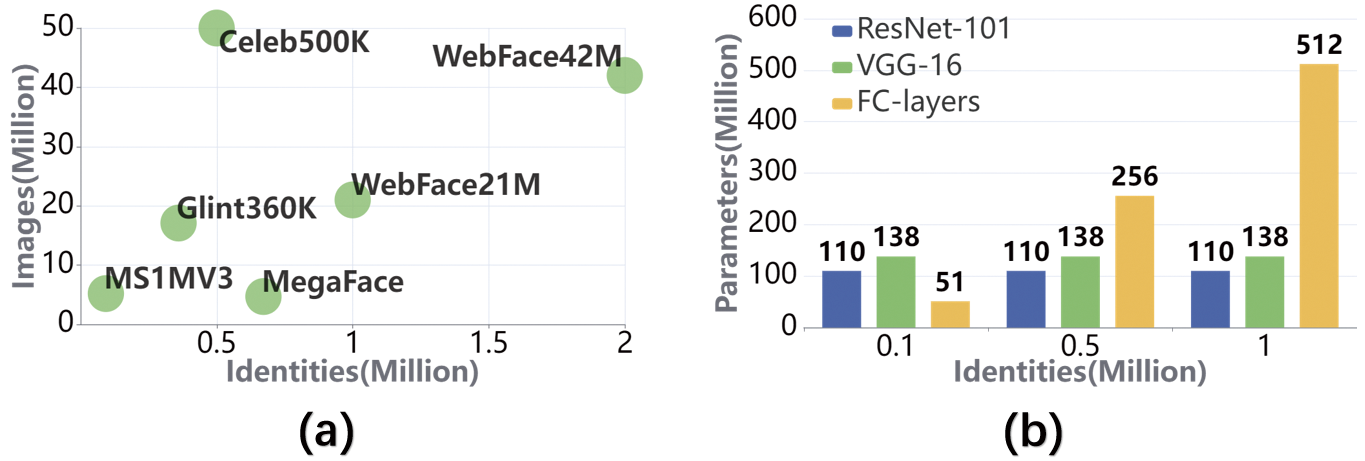} 
\caption{(a). The size of common large-scale face datasets. (b).The number of parameters in the backbone network and FC layer. The dimension of the FC layer $W = D \times N$ is related to the number of identities $N$, and $D$ is the dimension of features (in this case $D$ equals 512).}
	\label{fg:id_param}
\end{figure}

The model applied to FR can be split into two main parts, the feature extractor and the classification head. MobileNet \cite{howard2017mobilenets}, VGG \cite{DBLP:journals/corr/SimonyanZ14a}, ResNet \cite{DBLP:conf/cvpr/HeZRS16}, or other neural network backbones are often selected as the feature extractor, while the classification head is usually a fully-connected (FC) layer. The essential problem is that the size of large-scale datasets is huge, as demonstrated in Fig. \ref{fg:id_param}(a), and the dimensionality of the FC layer is linearly related to the number of identities as shown in Fig. \ref{fg:id_param}(b). In a large-scale FR scene, the parameters in the FC layer will be much higher than those in the backbone network. Therefore, when using large-scale datasets to train the model, hundreds of millions of parameters in the FC layer would consume considerable memory and time. In conclusion, the original FC layer used as the classification head consumes massive computing resources during training with large-scale datasets. 
\par
To address these issues, several state-of-the-art approaches have been proposed. Partial FC \cite{DBLP:conf/cvpr/AnDGFZ0L22} allocates the whole FC layer into multiple GPUs. It reduces the memory consumption on each single GPU but requires more GPUs to train the model, which is hardware-unfriendly. Virtual FC (VFC) \cite{DBLP:conf/cvpr/LiW021}, Dynamic Class Queue 
(DCQ) \cite{DBLP:conf/cvpr/0005XZFHLDL21}, and Faster Face Classification (F$^2$C) \cite{DBLP:conf/cvpr/WangWZZZWPSLY22} all create a new structure with fewer parameters to replace the FC layer. The main difference among them is the way they generate class centers. DCQ uses one image to generate the class center. F$^2$C designs two data loaders for generating class centers. VFC generates the class center with multiple images, but it considers the weight of each image to be equal.
\begin{figure*}[t]
\centering

\includegraphics[width=1.5\columnwidth]{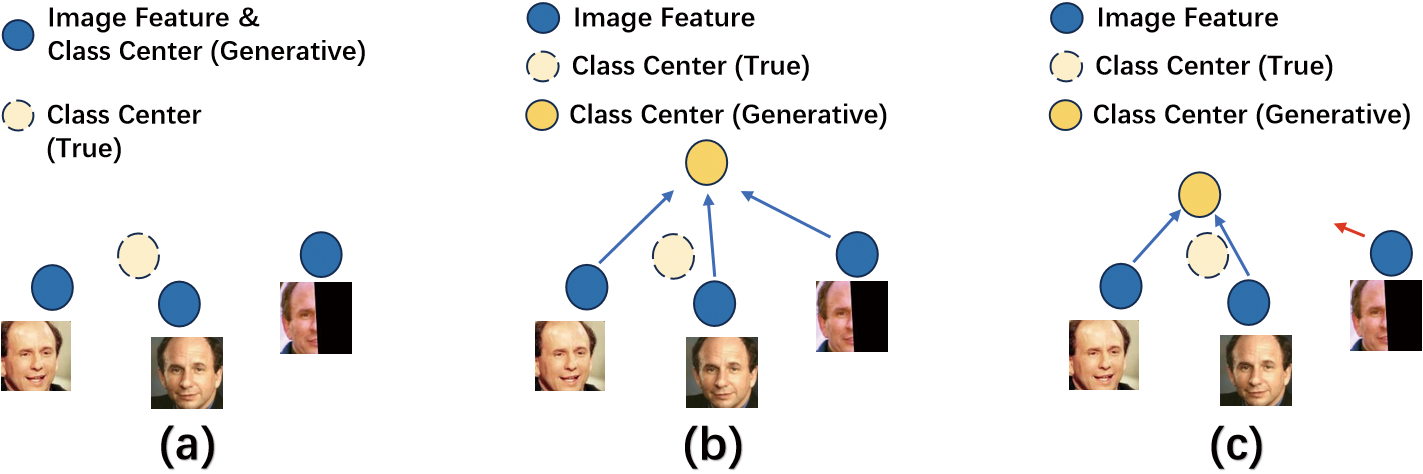}
	
    \caption{Comparison of different strategies for generating GCC: (a) Generating GCC with one image: The GCC from low-quality images may be dissimilar to TCC and other normal features. (b) Generating GCC with multiple images: The GCC is as similar as possible to the image features, but it may deviate further from TCC if the contribution of low-quality images is too large. (c) Generating GCC with multiple images based on attention weight: The contribution of low-quality images is decreased, thus the GCC is closer to TCC.}
\label{fg:GCC_with_different_image}
\end{figure*}
\par
Generally, each identity in the dataset corresponds to a unique true class center (TCC) in the feature space. The features extracted from the same identity images are distributed around the TCC. It could also be considered that the features and TCC are similar. However, the TCC is an abstract concept that cannot be represented in the real world. Thus, an approximate estimation of the TCC is commonly used to work as the TCC. When utilizing the original FC layer for FR model training, each class center in the FC is constantly updated based on stochastic gradient descent (SGD) or other optimization algorithms to get closer to the TCC. The updated class center is called the learned class center (LCC). Furthermore, since the feature extracted from the image closely resembles the corresponding identity TCC, we propose the generative class center (GCC) to represent this feature. Both GCC and LCC can function as the approximate estimation of TCC. The most important is to ensure that the estimated class center does not deviate heavily from the TCC. LCC is updated based on the images in the dataset, which fulfills this requirement. However, for large-scale datasets, the magnitude of LCCs is huge because the number of LCCs is equal to the number of identities. Therefore, updating all LCCs at each iteration will consume significant resources. For GCC, when generating the GCC from a single image, it is difficult to represent the TCC properly. Because each feature could be regarded as GCC, and the GCC from low-quality images may be dissimilar from TCC and other normal features as shown in Fig. \ref{fg:GCC_with_different_image}(a). Therefore generating the GCC with multiple images is a more rational approach. Since the GCC from multiple images ensures the similarity to image features, thus it better serves as TCC. However, large-scale datasets contain a variety of low-quality images. Thus it is crucial to control the contribution of these noisy images when generating the GCC. Otherwise, the GCC would deviate further from TCC, as illustrated in Fig. \ref{fg:GCC_with_different_image}(b). 

\par
In this work, we propose Attention FC (AttFC) to address the issues above. AttFC utilizes a Dynamic Class Container (DCC) to store the GCCs. DCC can be considered as a replacement for the FC, but DCC stores only a small subset of all class centers in the FC layer, and the one retained in the DCC is updated in each iteration. Hence the parameters of DCC are much smaller than FC, conserving training resources. Additionally, inspired by attention mechanism \cite{DBLP:conf/nips/VaswaniSPUJGKP17} and Momentum Contrast (MoCo) \cite{DBLP:conf/cvpr/He0WXG20}, we design an attention loader and two encoders (feature encoder and class encoder) to calculate the contribution of different images in generating the GCC. This contribution is related to attention, thus referred to as attention weight. By using the attention weight, the influence of low-quality images can be reduced, while enhancing the significance of high-quality ones, leading to GCC more closely resembling TCC, as illustrated in Fig. \ref{fg:GCC_with_different_image}(c). Additionally, we conduct extensive experiments to demonstrate that using AttFC can ensure the model's excellent performance while reducing training resources. The contributions of our method can be summarized as follows.
\begin{itemize}
  \item We propose AttFC to significantly decrease the training resources for the FR model on large-scale datasets, including computational time and memory space, leading to a lighter hardware resource requirement. At the same time, the model can still maintain state-of-the-art performance.
  \item We propose using GCC to work as TCC properly and implementing DCC to replace the FC for storing the GCCs. DCC only stores a small part of class centers in the FC layer, significantly reducing the parameter count.
 \item We employ the attention mechanism to adjust the contribution of different quality images when generating the GCC, leading to GCC more closely resembling TCC.
\end{itemize}

\section{Related Work}
\label{sc:RelatedWork}
Over the years, many remarkable research results have been achieved in the field of face recognition. The Convolutional Neural Networks \cite{DBLP:journals/corr/SunLWT15} helps solve the problem of face recognition. Then the Large-Margin Softmax Loss \cite{DBLP:conf/icml/LiuWYY16} explicitly encourages intra-class compactness and inter-class separability between learned features. CosFace \cite{DBLP:conf/cvpr/WangWZJGZL018} and ArcFace \cite{DBLP:conf/cvpr/DengGXZ19} cut into from the angular space, further increasing the angle between different classes and bringing the same class closer. AnchorFace \cite{DBLP:conf/aaai/LiuQWL22} bridges the gap between the training and evaluation process for FR. Random Temperature Scaling \cite{DBLP:conf/aaai/ShangHSLLSSXQ23} can adjust the learning strength of clean and noisy samples to enhance the model performance. Numerous previous works \cite{DBLP:conf/iccvw/AnZGXZFWQZZF21,DBLP:conf/cvpr/ZhuHDY0CZYLD021,DBLP:conf/cvpr/0005XZFHLDL21,DBLP:journals/tip/HuPYHV18} have shown that 
large-scale datasets can enhance model performance. However, larger datasets mean more identities, which results in more parameters in the FC layer and more training resources. To solve this problem, a variety of methods are proposed. Partial FC \cite{DBLP:conf/cvpr/AnDGFZ0L22} decreases train consumption by splitting the FC layer. VFC \cite{DBLP:conf/cvpr/LiW021} uses a grouping and re-grouping strategy to make the identities in a group share an anchor and combine all anchors to form a virtual FC layer. DCQ \cite{DBLP:conf/cvpr/0005XZFHLDL21} uses a moving average updated siamese network \cite{DBLP:conf/eccv/DuSLWL0020} to dynamically generate the class center so that displace the FC layer. It reduces the computation overhead and also solves the long tail distribution problem. F$^2$C \cite{DBLP:conf/cvpr/WangWZZZWPSLY22} design a dual data loader including identity-based and instance-based loaders to improve the update efficiency of the class center.
\par

\section{Proposed Approaches}
\label{sc:pa}
In this section, we will review the traditional FC layer and analyze its drawbacks on large-scale datasets. After that, Attention FC, the objective of which is solving these problems, will be introduced in detail.  
\subsection{FC Layer in Large-Scale Datasets}
\label{ssc:FCL}
While training a FR model, the softmax loss is usually used as the loss function:
\begin{equation}
\label{eq:softmaxloss}
        L=-\frac{1}{B}
        \log 
	\sum_{i=1}^{B}
        \frac{e^{\langle w^+,f_i \rangle}}
        {e^{\langle w^+,f_i\rangle}+\sum\limits_{j\in M} e^{\langle w^-_j, f_i\rangle}}
\end{equation}
where $B$ represents the mini-batch size, $f_i \in R^D$ denotes the $i$-th image feature in the batch, while $w^+ \in R^D$ is the positive class center corresponding to $f_i$. $M$ refers to the set of other negative classes and the size is 
$\left \|M\right\|=N-1$, in which $N$ is the number of identities, $w^-_j \in R^D$ is the $j$-th negative class center. $\langle \cdot,\cdot \rangle$ indicates the similarity function, typically represented by the inner product operation. To increase the angular margin, alternative similarity functions such as ArcFace \cite{DBLP:conf/cvpr/DengGXZ19} could be employed, which is defined as:
\begin{equation}
\label{eq:logit}
        \langle w,f_i\rangle =
        \left\{\begin{matrix}
          s(\cos({\theta + m})),w = w^+ \\
          s(\cos({\theta})),w \neq w^+
        \end{matrix}\right.
\end{equation}
where $\theta$ is the angle between the class center $w$ and image feature $f_i$, $s$ is the feature scale, $m$ is the margin penalty.
\par
For the feature $f_i$, the output of similarity function $\langle \cdot,\cdot \rangle$ can be interpreted as the degree of similarity between the image feature and each class center. Mapping by softmax, this degree corresponds to the classification probability that the image belongs to each class, in which $p^+$ is the probability that $f_i$ belongs to the corresponding positive class (Eq. \eqref{eq:w_positive}). Relatively, $p^-_j$ refers the probability that $f_i$ belongs to the $j$-th negative class in $M$ (Eq. \eqref{eq:w_negtive}).
\begin{equation}
	\label{eq:w_positive}
        p^+ = \frac{e^{\langle w^+,f_i \rangle}}
          {e^{\langle w^+,f_i\rangle}+\sum\limits_{j\in M} e^{\langle w^-_j, f_i\rangle}}
\end{equation}
\begin{equation}
	\label{eq:w_negtive}
        p^-_j = \frac{e^{\langle w^-_j,f_i \rangle}}
          {e^{\langle w^+,f_i\rangle}+\sum\limits_{j\in M} e^{\langle w^-_j, f_i\rangle}} 
\end{equation}
in this case  Eq. \eqref{eq:softmaxloss} can also rewritten as Eq \eqref{eq:positive_negative_loss} for simplicity.

\begin{equation}
	\label{eq:positive_negative_loss}
        L= -\frac{1}{B}
        \log 
	\sum_{i=1}^{B}p^+_i
\end{equation}
where $p^+_i$ represents the classification probability that the $i$-th image belongs to the positive class. Then according to the backpropagation (BP) algorithm, $\frac{\partial L}{\partial f_i} $and $\frac{\partial L}{\partial w_i}$ are defined as:
\begin{equation}
	\label{eq:partial_f}
        \frac{\partial L}{\partial f_i} = -(1-p^+) w^+ + \sum_{j\in M}p_{j}^-w_{j}^-
\end{equation}

\begin{equation}
	\label{eq:partial_W}
        \frac{\partial L}{\partial w_i} = -\sum_{x\in I^+}(1-p_{x}^+) f_x + \sum_{y\in I^-}p_{y}^-f_y
\end{equation}
where $I^+$ denotes all the images belonging to $i$-th class in the dataset, while $I^-$ represents those of other classes. 
\par
In a dataset consisting of $N$ identities, each identity corresponds to a positive class and $N-$1 negative class. Assuming the feature dimension is represented by $D$, the size of the FC layer can be denoted as $W = D\times N$. During training the model will converge that the feature $f_i$ becomes closer to the positive class center $w^+$, and moves further away from the other negative class center $w^-$. It is the large-scale datasets that improve the recognition ability of the FR model. Still, since the size of the FC layer is related to the number of identities $N$, large-scale datasets also result in a large number of parameters in the FC layer, and the computational cost of updating $N$ LCCs at each iteration is substantial. A straightforward method to conserve computational resources is directly reducing the number of class centers to $S$ ($S\ll{N}$). However, this approach would introduce new issues. To be specific, the images belonging to other $N-S$ identities can not be classified into a certain class, which means that the model cannot calculate the loss by Eq. \eqref{eq:softmaxloss}. Accordingly,  the model could not be updated with the BP algorithm.
\begin{figure*}[t]
    \centering
\includegraphics[width=1.85\columnwidth]{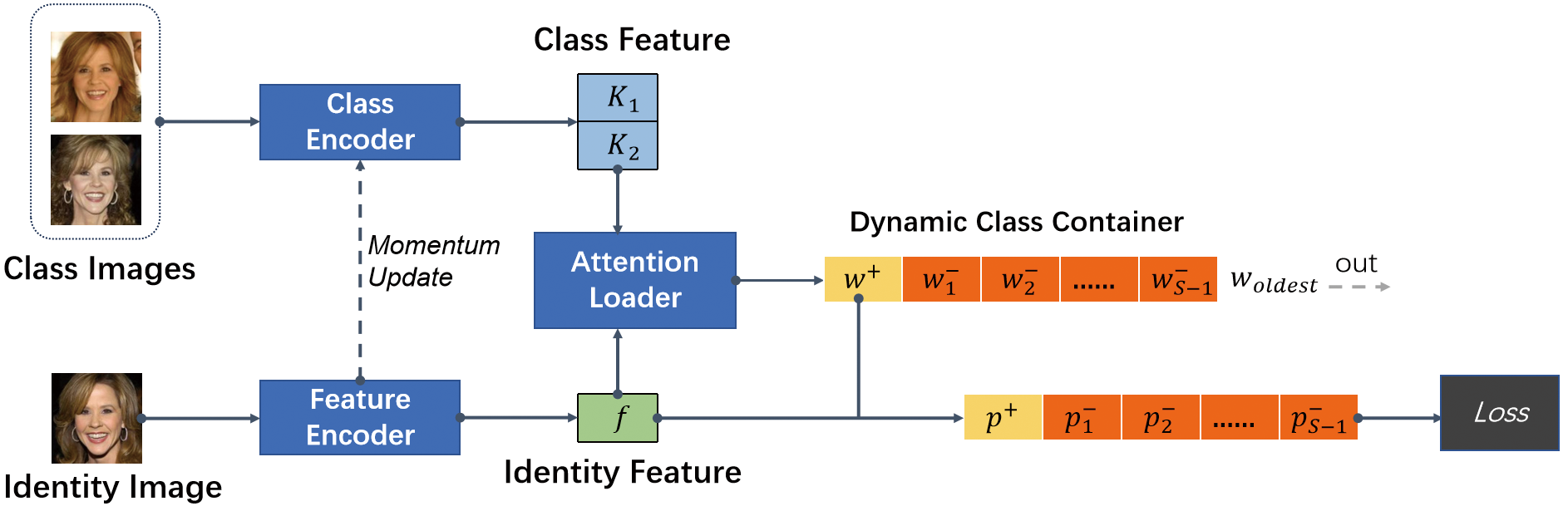}
    \caption{The architecture of AttFC. AttFC uses a feature encoder and class encoder to obtain the identity feature and class features respectively. The attention loader will generate the GCC with these features and store it in DCC. DCC works as an FC layer for calculating the loss. Finally, the feature encoder is updated with SGD, while the class encoder will be momentum updated by the feature encoder.}
    \label{fg:dcc}
\end{figure*} 

\subsection{Attention FC}
\label{ssc:DCC}
To solve the above issues, we propose AttFC to significantly reduce computational resources, the architecture of AttFC as illustrated in Fig. \ref{fg:dcc}. 
The AttFC differs from the standard FR model as it includes two backbone networks: the feature encoder and the class encoder, both networks have the same structure. During the data loading phase, two types of images are input to the feature encoder and class encoder, respectively. One is the identity image, which is input into the feature encoder to extract the identity feature $f \in R^D$. The other is the class image, which is in the same class as the identity image, and for each identity image, $k$ additional class images are input into the class encoder. The output of the class encoder is referred to as the class features $K \in R^{k\times D}$, and $K_i$ is the feature from $i$-th class image. The identity feature and class features are then input into the attention loader to generate the GCC. The GCC will be stored in the Dynamic Class Container (DCC) and replace the oldest GCC previously generated. DCC is a first-in-first-out (FIFO) queue with a capacity of $S$, in which the queue element is the class center $w \in R^D$. So DCC can also be viewed as an FC layer with the size $W_{DCC} = D \times S$ and used to calculate the loss with the identity feature $f$. Finally, according to the BP algorithm, the feature encoder is updated based on SGD, while the class encoder which has an identical structure to the feature encoder, is momentum-updated based on the feature encoder.
\par
When using AttFC to train the FR model, it is equivalent to using a subset of the face dataset as the training dataset. The size of the subset is fixed and far smaller than the origin dataset, and the images of the subset are randomly selected from the origin dataset in each iteration. Hence the number of identities in the dataset becomes smaller, resulting in the decrease of parameters in the FC layer. The key to this idea is, since a face dataset only contains a small portion of all the faces in the real world, it's feasible to use a smaller subset of the face dataset for model training\cite{DBLP:conf/iccvw/AnZGXZFWQZZF21}. This means reducing the number of identities, which may weaken the discriminative capability of the FR model. But benefiting from the strong learning ability of DNNs and sufficient samples of large-scale datasets, using AttFC could also maintain the model performance. It is worth noting that, AttFC only replaces the FC layer, and then introduces additional model components such as the class encoder and attention loader based on the standard FR model, which means AttFC can be easily integrated with other methods.
\par
\textbf{\normalsize Attention Loader.}
The attention mechanism can be understood as gathering crucial information from a large amount of data and focusing on it. In simple terms, it is a calculation of weighted sums, in which the weight depends on the data's level of importance. For a given identity, its corresponding TCC contains substantial crucial information about this identity. If the GCC of this identity also contains more important information, it would be more similar to the TCC. While generating GCC with multiple images, the attention mechanism can be employed to estimate the weight of each image. By using these weights to generate the GCC with multiple images, the GCC would contain more important information and be closer to the TCC.
\par
In AttFC, an attention loader is designed to generate the GCC based on the attention mechanism. There are various forms of attention, such as additive attention \cite{DBLP:journals/corr/BahdanauCB14}, scaled dot-product attention \cite{DBLP:conf/nips/VaswaniSPUJGKP17}, etc. Inspired by these methods, the attention loader determines the importance of each class feature by calculating the cosine similarity between $f$ and $K_i$, and then using the softmax function to convert the similarity into the attention weight $\alpha \in R^{k}$. The computation of $\alpha$ can be summarized as Eq. \eqref{eq:attention weight}.
\begin{equation}
	\label{eq:attention weight}
        \alpha= softmax(similarity(f,K))
\end{equation}
where $similarity(f,K) \in R^k$ denotes to the cosine similarity between $f$ and each $K_i$. Then, each $K_i \in R^D$ is multiplied by its corresponding $\alpha_i$ for a weighted sum, resulting in the positive class center $w^+ \in R^D$ as shown in Eq. \eqref{eq:w_pos}.
\begin{equation}
	\label{eq:w_pos}
        w^+= \sum_{i=1}^{k}\alpha_i K_i
\end{equation}
\par
It is important to note that two features with high similarity often contain repetitive or similar information, which can serve as important information to represent the identity. Therefore, we design the attention loader to calculate the attention weight by cosine similarity. If a class feature has a higher similarity to the identity feature, it signifies that the class feature contains more important information, then its attention weight is increased. Compared with calculating $\alpha$ with the attention loader, another way is to set $\alpha$ as a constant value. Assume that $\alpha_i=1/k$, which means that each class image is equally important for generating GCC. However, in large-scale datasets, there may be a variety of low-quality images, such as missing edges, blurred, low resolution, and so on. Class features extracted from these problematic images are far away from TCC in the feature space. When using the constant weight for these features, the GCC may be far away from TCC. But if calculating the attention weight with the attention loader, the weights for these low-quality features would decrease, because the similarity between the low-quality feature and the identity feature is usually not high. So the attention loader ensures that GCC does not deviate from TCC.
\par
\begin{figure}[t]
\centering
\includegraphics[width=0.95\columnwidth]{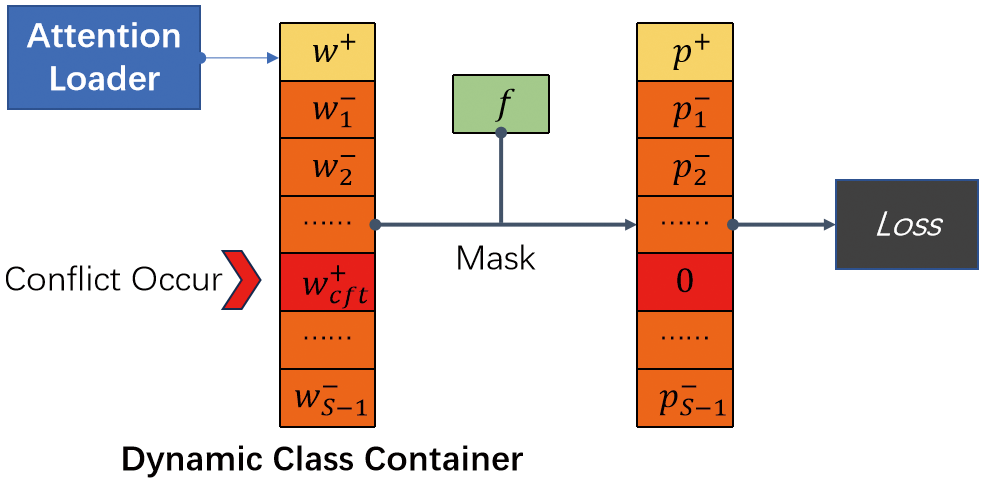}
     \hfil
    \caption{
    If the class center conflict occurs during training, it is necessary to mask the superfluous $w^+_{cft}$, ensuring that the probability of $f$ belonging to the class center $w^+_{cft}$ is 0.
    }
    \label{fg:mask}
\end{figure}
\textbf{\normalsize Dynamic Class Container.}
 A straightforward method to save computational resources is to decrease the number of class centers in the FC layer from $N$ to $S$, where $S$ is much smaller than $N$. In AttFC, a Dynamic Class Container is designed to replace the FC layer and store $S$ class centers. From the perspective of data structure, DCC is a first-in-first-out (FIFO) queue with a capacity of $S$, each queue element is a GCC generated from the attention loader. So the overall size of DCC is $W_{DCC}= D \times S$. At the beginning of model training, DCC is initialized with random values from a normal distribution. Within each iteration, the attention loader generates a batch-size number of GCCs, which are then input into the DCC. Subsequently, the oldest batch of GCCs is out of the queue. The dynamic updating process of DCC 
guarantees that, even though DCC only stores $S$ class centers, there will always be a corresponding $w^+$ for each identity feature $f$. DCC addresses the issue that the FC layer is unable to classify an image belonging to other $N-S$ identities to a certain class.
\par
 \begin{algorithm}[t]
\caption{Mask}
\label{alg:mask}
\textbf{Input}: Dynamic class container {$W_{DCC}$}, identity feature {$f$}\\
\textbf{Output}: Classification probability {$p$}\\     
\begin{algorithmic}[1]
\STATE$logits = \langle W_{DCC},f \rangle \in R^S$.
\IF{$w^+_{cft}$ exist}
\STATE $index =$GET-INDEX($w^+_{cft}$)
\STATE $logits[index] = -\infty$
\ENDIF
\STATE $p =$ Softmax$(logtis)\in R^S$
\STATE \textbf{return} $p$
\end{algorithmic}        
\end{algorithm}
It should be emphasized that theoretically, there exists only one $w^+$ for an identity feature. However, since the FIFO update strategy is employed for DCC, and the training datasets are typically input to the model in batches, it is inevitable to sample images with the same label in the same or consecutive batches. Suppose the $i$-th feature $f$ in the current batch belongs to the $j$-th class, it would correspond to generate a $w^+$. If there already exists one or more $w^+_{cft}$ in the DCC that were generated by other features belonging to $j$-th class, then a class center conflict occurs. This conflict means multiple $w^+$ for one identity feature, that is, one image belonging to multiple identities. 
The conflict would cause the model to learn in the wrong direction, which would affect the model's discriminative ability. Therefore, it is necessary to mask \cite{DBLP:conf/cvpr/0005XZFHLDL21} the superfluous $w^+_{cft}$ while computing the classification probability as illustrated in Fig. \ref{fg:mask}, and the mask algorithm refers to the Alg. \ref{alg:mask}. Then the loss of AttFC can be summarized as follows:
\begin{equation}
	\label{eq:dcc_loss}
        L_{AttFC}=-\frac{1}{B}
        \log 
	\sum_{i=1}^{B}
        \frac{e^{\langle w^+,f_i \rangle}}
        {e^{\langle w^+,f_i\rangle}+\sum\limits_{j\in M_S} e^{\langle w^-_j, f_i\rangle}}
\end{equation}
compared with the regular softmax loss (Eq. \eqref{eq:softmaxloss}), 
because AttFC uses DCC to replace the FC layer, for the feature $f_i$, the negative classes set $M_S$ denotes all the class centers in DCC except for $w^+$ and $w^+_{cft}$. As a result, the size of $M_S$ is $\lvert\lvert M_S \rvert\rvert = S - 1 - \lvert\lvert w^+_{cft} \rvert\rvert$, where $\lvert\lvert w^+_{cft} \rvert\rvert$ represents the number of $w^+_{cft}$ corresponding to $f_i$. $\lvert\lvert M_S \rvert\rvert$ is significantly smaller than $N$, leading to a considerable savings in computational resources.
\par
In a complete training iteration, the loss is calculated at the forward propagation (FP). The next step involves computing the gradient and proceeding with the backpropagation (BP). Since DCC stores GCC rather than LCC, there is no necessity to calculate $\frac{\partial L}{\partial w_i}$. This showcases that AttFC can effectively conserve computing resources in both FP and BP. For the update of feature encoder, $\frac{\partial L_{AttFC}}{\partial f_i}$ can still be calculated based on Eq. \eqref{eq:partial_f}, the only change is the set of negative classes from $M$ becomes to $M_S$:
\begin{equation}
	\label{eq:partial_f_attFC}
        \frac{\partial L_{AttFC}}{\partial f_i} = -(1-p^+) w^+ + \sum_{j\in M_S}p_{j}^-w_{j}^-
\end{equation}
for the class encoder, its network structure is equivalent to the feature encoder, and the momentum update \cite{DBLP:conf/cvpr/He0WXG20} is used for it, as demonstrated in Eq. \eqref{eq:moco_update}. 
\begin{equation}
	\label{eq:moco_update}
        \theta_{ce} = \gamma \theta_{ce} + (1-\gamma) \theta_{fe}
\end{equation}
where $\theta_{ce}$ and $\theta_{fe}$ represent the parameters of the class encoder and feature encoder, respectively. $\gamma$ is the momentum factor. The momentum update helps maintain the consistency of the class encoder, which ensures that the update of the class encoder is conservative. Even after several iterations of training, the class feature used for generating the GCC is still similar to the previous one, which prevents feature drift.
\section{Experiments}
\label{sc:exp}
\subsection{Datasets}
In this paper, we thoroughly evaluate the AttFC with two large-scale training datasets. The first is MS1MV3 \cite{DBLP:conf/eccv/GuoZHHG16}, which comprises 93K identities and 5M images. The second is the subset of WebFace42M \cite{DBLP:conf/cvpr/ZhuHDY0CZYLD021}, including WebFace21M which consists of 1M identities and 21M images, as well as smaller subsets like WebFace8M and WebFace4M, which contain 400K and 200K identities, respectively. For model testing, we demonstrated the performance of AttFC on the standard benchmarks including LFW \cite{Huang2008LabeledFI}, CFP-FP \cite{DBLP:conf/wacv/SenguptaCCPCJ16}, AgeDB \cite{DBLP:conf/cvpr/MoschoglouPSDKZ17}, IJB-B \cite{DBLP:conf/cvpr/WhitelamTBMAMKJ17}, IJB-C \cite{DBLP:conf/icb/MazeADKMO0NACG18}. For IJB-B and IJB-C, TAR@FAR = 1e-4 is used.
\par
\subsection{Configurations} 
All experimental code in this paper is implemented based on Pytorch and runs on a single GeForce RTX 4090 24G GPU card. ResNet \cite{DBLP:conf/cvpr/HeZRS16} is used as the backbone of AttFC. For quick experiments, the training epoch is 5 and the batch size is set to 384 by default. SGD with a momentum of 0.9 and a weight decay of 0.0005 is employed for model updating. The learning rate is initialized to 0.1, and cosine annealing decay \cite{DBLP:conf/iclr/LoshchilovH17} is employed for the learning rate. The feature dimension is set to 512. In the loss function, the scale is 64 and the margin is 0.5 for ArcFace \cite{DBLP:conf/cvpr/DengGXZ19}. For the AttFC, the capacity of DCC $S$ is calculated by $r\times N$, where $N$ represents the number of identities and $r$ is the size ratio, a hyper-parameter with the default value of 0.3. The number of class images $k$ is set to 2 to balance model performance and training time. For the momentum factor $\gamma$, the default value 0.999 in MoCo \cite{DBLP:conf/cvpr/He0WXG20} is used. 

\begin{table}[t]
\centering
\setlength{\tabcolsep}{1.7mm}
{
\begin{tabular}{c|c|c|c|c|c|c} 
        \hline
        Batch & Model & LFW & CFP & AgeDB & IJB-B & IJB-C \\ [0.5ex] 
         \hline\hline
         \multirow{2}{*}{256} & FC & 99.72 & 97.73 & 97.82 & 94.57 &96.05  
         \\
           ~ & AttFC & 99.77 & 97.64 & 97.58 & 93.88 &  95.43
         \\
         \hline
          \multirow{2}{*}{384} & FC & 99.77 & 98.03 & 97.73 & 94.76 & 96.18 
         \\
           ~ & AttFC &99.73 & 97.86 & 97.50 & 94.20 & 95.78
         \\
         \hline
        \end{tabular}
}
    \caption{Model accuracy comparison between FC and AttFC on MS1MV3 datasets with different batch sizes. ResNet50 is used here.} 
\label{tab:FC_and_AttFC_EP5}
\end{table}
\subsection{Exploratory Experiment}
\hspace{1em}\textbf{Contrast experiment with FC.}
To assess the impact of using DCC to store only a small portion of all class centers on the model performance, we compared AttFC and FC with the MS1MV3 dataset. Although MS1MV3 is quite large, compared to other larger datasets shown in Fig. \ref{fg:id_param}(a), it can still be trained with FC on a single GPU, which is suitable as the baseline in the experiments. The results as shown in Tab. \ref{tab:FC_and_AttFC_EP5}, with the same configuration settings, the maximum decrease on each test set is less than 1$\%$. However, the number of parameters in AttFC is reduced to 30$\%$ of FC, while AttFC performs comparably to FC. In addition, it can be observed that the model performs better as the batch size increases. According to the Law of Large Numbers in statistics, the optimal method for model training is to input all data samples into the model within each iteration, if we possess abundant computational resources.
But in practice, the training is typically done with Mini-Batch Gradient Descent (MBGD) \cite{DBLP:journals/corr/Ruder16} for conserving resources. The model's update is based on a batch-size number of samples. A larger batch size allows the model to consider more perspectives, enhancing the generalization ability and leading to better performance. For AttFC, a larger batch size also implies that more GCC will be updated in each iteration. These newly generated GCCs tend to be closer to the TCC than the previously generated ones, thereby improving model performance.

\begin{table}[t]
\centering
\setlength{\tabcolsep}{1.3mm}
    \begin{tabular}{c|c|c|c|c|c} 
        \hline
        Model & $W$-size & LFW & CFP & AgeDB & IJB-C \\ [0.5ex] 
         \hline\hline
         \multicolumn{6}{c}{MS1MV3} \\
         \hline
         FC & 93431 & 99.77 & 98.03 & 97.73  & 96.18 \\
         AttFC-0.1 & 9216 & 99.70 & 97.63 & 97.45  & 95.20 \\
         AttFC-0.2 & 18432 & 99.67 & 97.64 & 97.55  & 95.43 \\
         AttFC-0.3 & 27648 & 99.73 & 97.86 & 97.50  & 95.78 \\
         \hline\hline
          \multicolumn{6}{c}{WebFace4M} \\
        \hline
         FC & 205990 & 99.75 & 98.64 & 97.53 & 96.27 \\
         AttFC-0.1 & 20352 & 99.60 & 98.36 & 97.05 & 95.68 \\
         AttFC-0.3 & 61056 & 99.65 & 98.43 & 97.00 & 95.86 \\
        \hline\hline
          \multicolumn{6}{c}{WebFace8M} \\
        \hline
         FC & 411980 & 99.77 & 98.90 & 97.50  & 96.64 \\
         AttFC-0.1 & 41088 & 99.67 & 98.74 & 97.33 & 96.32 \\
         AttFC-0.3 & 123264 & 99.75 & 98.77 & 97.53  & 96.45 \\
         \hline\hline
         \multicolumn{6}{c}{WebFace21M} \\
        \hline
         FC & 1029950 &\multicolumn{4}{c}{Out of Memory}\\
         Partial FC-0.3 & 308736 &\multicolumn{4}{c}{Out of Memory}\\
         AttFC-0.3 & 308736 & 99.82 & 99.07 & 98.02 & 97.01 \\
         \hline
        \end{tabular}
    \caption{Performance comparison with different datasets and size ratio for AttFC. The FC model is used as the baseline and ResNet50 is used here. AttFC-0.3 refers to the AttFC model with a size ratio of 0.3. Similarly, Partial FC-0.3 represents the same $W$-size as AttFC-0.3.}
\label{tab:AttFC_diff_datasets_size}
\end{table}

\par
\textbf{\normalsize AttFC with different datasets and size ratio.}
In Tab. \ref{tab:AttFC_diff_datasets_size}, we present the influence of various datasets and container sizes on the performance of AttFC with ResNet50. The "$W$-size" column represents the number of class centers in the classification head, which corresponds to the number of identities $N$ for FC and the capacity of DCC $S$ for AttFC. The table demonstrates that as the size ratio increases, the performance of AttFC improves. A larger size ratio leads to more class centers in DCC, which augment the model's classification ability. Employing larger datasets for model training could also improve the accuracy since the capacity of DCC is increased, too. Compare FC and AttFC trained with different datasets, for example, the FC trained with WF4M (with a $W$-size of 205,990) versus the AttFC-0.3 trained with WF8M (with a $W$-size of 123,264), and the FC trained with WF8M (with a $W$-size of 411,980) versus the AttFC-0.3 trained with WF21M (with a $W$-size of 308,736), it is evident that despite each AttFC model having a smaller $W$-size than the respective FC model, they also demonstrate comparable or even superior performance. For some huge-scale datasets, such as WebFace21M which consists of 1M identities, the model training with FC was interrupted by an out-of-memory(OOM) error, while the training with AttFC proceeded normally. Similarly, Partial FC, a leading state-of-the-art method for large-scale face recognition, also encountered OOM issues under the same configuration as AttFC. This demonstrates that AttFC can significantly save computational resources while maintaining the model's performance.
\begin{table}[!t]
\centering
    \begin{tabular}{c|c|c|c|c|c|c} 
        \hline
        \rule{0pt}{8pt}
        k & Mem &Speed &  LFW & CFP & AgeDB & IJB-C \\ [0.5ex] 
         \hline\hline
         2 & 15.5& 788 & \textbf{99.77} & 97.64 & 97.58  & 95.43  \\
         3 & 18.3 & 638 & 99.72 & \textbf{97.86} & \textbf{97.62}  & 95.49 \\
         4 & 21.0 & 545 & 99.63 & 97.67 & 97.60  & \textbf{95.56} \\
         5 & 23.2 & 470 & 99.70 & 97.70 & 97.53 & 95.50 \\
         \hline
        \end{tabular}
    \caption{The memory consumption and training time of AttFC on MS1MV3 with different $k$ value. Batch size is set to 256 and ResNet50 is used here. Mem refers to the GPU memory consumed in GB. Speed denotes throughput in samples/second.
    }
\label{tab:K_value}
\end{table}

\begin{table}[!t]
\centering
    \begin{tabular}{c|c|c|c|c} 
        \hline
  Strategies  & LFW & CFP & AgeDB  & IJB-C \\ [0.5ex] 
         \hline\hline
         Single image & 99.72 & 97.61 & 97.40 & 94.93  \\
         \hline
         Multiple images\\
         Constant weight & 99.70 & 97.39 & \textbf{97.67} & 95.31  \\
        Attention weight & \textbf{99.73} & \textbf{97.86} & 97.50 & \textbf{95.78} \\
         \hline
        \end{tabular}
    \caption{Performance comparison in generating GCC with different strategies. The training dataset is MS1MV3 and ResNet50 is used here.}
\label{tab:attention_constant}
\end{table}

\textbf{\normalsize AttFC under a different number of class images.}
For AttFC, GCC is generated from multiple class images. In Tab. \ref{tab:K_value}, we demonstrate the influence of different numbers of class images $k$ on model performance. The training dataset is MS1MV3, and RestNet50 is used here. The table illustrates that as $k$ increases, the model performance tends to improve. This occurs because a higher $k$ means that the GCC is associated with more class features, leading it to be closer to TCC.  However, a larger $k$ also requires additional resources to process the extra images, and it is worth noting that the accuracy improvement is minimal when $k > 3$. Consequently, considering the trade-off between model performance and computational resources, we set the default value of $k$ to 2.
\par
\textbf{\normalsize Comparison of generating GCC with different strategies.}
In Tab. \ref{tab:attention_constant}, we conduct comparative experiments to generate GCC with different strategies, which include generating GCC with a single image, with multiple images based on constant weight, and with multiple images based on attention weight. The backbone is ResNet50 and the training dataset is MS1MV3. 
Constant weight refers to setting the $\alpha_i$ in Eq. \eqref{eq:w_pos} to a constant value. In this case, $\alpha_i$ is set to $1/k$, implying that each image contributes equally when generating the GCC. 
The table shows that generating GCC with multiple images is better than using just a single image. When using a single image to generate GCC, the class center is difficult to serve as TCC properly. This is because the GCC from a low-quality image may be dissimilar from TCC, making accurate classification challenging. Furthermore, the model employing attention weight tends to outperform the one with constant weight. Since not every image is equally valuable for generating GCC, compared to those from low-quality images, the class features extracted from high-quality images are more important. By utilizing attention weight, the contribution of these high-quality images increases, while low-quality images have a lower contribution when generating GCC. As a result, the GCC is more closely with TCC, then improving the model performance.

\begin{table}[!t]
\centering
\setlength{\tabcolsep}{1.4mm}
\begin{tabular}{c|c|c|c|c|c} 
        \hline
        Method & CALFW & CPLFW & CFP-FP & IJB-C &Avg \\ [0.5ex] 
         \hline\hline
        Partial FC & 95.90 & 92.17 & 97.83 & 95.37 & 95.32\\
         VFC  & 91.93 & 79.00 & 95.77  & 70.12 & 84.21 \\
         DCQ  & 95.38 & 88.92 & 98.16 & 92.96 & 93.86 \\
        F$^2$C  & 95.25 & 89.38 & 98.25 & 92.31 & 93.80 \\
         \hline\hline
        AttFC  & 95.77 & 92.35  & 98.01 & 94.76 & 95.22 \\
         \hline
\end{tabular}

    \caption{Performance comparison between AttFC and state-of-the-art methods. To make a fair comparison, Partial FC, VFC, DCQ, F$^2$C, and AttFC only set the number of class centers in the classification head to 1$\%$ of identities of MS1M for training.}
\label{tab:SOTA_v2}
\end{table}

\textbf{Contrast experiment with SOTA methods.}
To make a fair comparison with other SOTA methods for large-scale face recognition, we referred to the experimental configurations in F$^2$C \cite{DBLP:conf/cvpr/AnDGFZ0L22} and trained AttFC with a size ratio 0.01 on MS1MV3. The training epoch is set to 20 and ResNet50 is used here. The result is presented in Tab. \ref{tab:SOTA_v2}, which indicates that AttFC outperforms most of the SOTA methods for large-scale face recognition. Our primary aim in proposing AttFC is not to achieve the highest accuracy, but rather to minimize computational resources and achieve a great balance between model performance and resource consumption. Some of these SOTA methods, such as Partial FC, would encounter an OOM error (as shown in Tab. \ref{tab:AttFC_diff_datasets_size}) when training on large-scale datasets with a single GPU, while AttFC can proceed normally. Moreover, AttFC only replaces the FC layer of the entire FR model, so it is convenient to combine with other methods. This shows AttFC's excellent scalability and potential for further development.
\section{Conclusion}
\label{sc:con}
In this paper, we propose AttFC for the training of the face recognition model with large-scale datasets. AttFC introduces the attention mechanism to generate the GCC and further utilizes the DCC to store a small number of class centers, which reduces the multitude of model parameters. With theoretical analysis and experimental validation, we demonstrate that AttFC can maintain high model performance while saving substantial computational resources.
\par
\textbf{\normalsize Future Improvement.} AttFC relies on multiple images of the same identity to generate the GCC. In future work, we will explore how to generate both features and the corresponding class center with just one image, which could further conserve computational resources.

\bibliography{AttFC}

\section{Reproducibility Checklist}
This paper:
\begin{itemize}
\item Includes a conceptual outline and/or pseudocode description of AI methods introduced (yes)
\item Clearly delineates statements that are opinions, hypothesis, and speculation from objective facts and results (yes)
\item Provides well marked pedagogical references for less-familiare readers to gain background necessary to replicate the paper (yes)
\end{itemize}
Does this paper make theoretical contributions? (yes)

If yes, please complete the list below.
\begin{itemize}
    \item All assumptions and restrictions are stated clearly and formally. (yes)
    \item All novel claims are stated formally (e.g., in theorem statements). (yes)
    \item Proofs of all novel claims are included. (yes)
    \item Proof sketches or intuitions are given for complex and/or novel results. (yes)
    \item Appropriate citations to theoretical tools used are given. (yes)
    \item All theoretical claims are demonstrated empirically to hold. (yes)
    \item All experimental code used to eliminate or disprove claims is included. (yes)
\end{itemize}
Does this paper rely on one or more datasets? (yes)

If yes, please complete the list below.
\begin{itemize}
    \item A motivation is given for why the experiments are conducted on the selected datasets (yes)
    \item All novel datasets introduced in this paper are included in a data appendix. (NA)
    \item All novel datasets introduced in this paper will be made publicly available upon publication of the paper with a license that allows free usage for research purposes. (NA)
    \item All datasets drawn from the existing literature (potentially including authors’ own previously published work) are accompanied by appropriate citations. (yes)
    \item All datasets drawn from the existing literature (potentially including authors’ own previously published work) are publicly available. (yes)
    \item All datasets that are not publicly available are described in detail, with explanation why publicly available alternatives are not scientifically satisficing. (NA)
\end{itemize}
Does this paper include computational experiments? (yes)

If yes, please complete the list below.
\begin{itemize}
    \item Any code required for pre-processing data is included in the appendix. (yes).
    \item All source code required for conducting and analyzing the experiments is included in a code appendix. (yes)
    \item All source code required for conducting and analyzing the experiments will be made publicly available upon publication of the paper with a license that allows free usage for research purposes. (yes)
    \item All source code implementing new methods have comments detailing the implementation, with references to the paper where each step comes from (partial)
    \item If an algorithm depends on randomness, then the method used for setting seeds is described in a way sufficient to allow replication of results. (NA)
    \item This paper specifies the computing infrastructure used for running experiments (hardware and software), including GPU/CPU models; amount of memory; operating system; names and versions of relevant software libraries and frameworks. (partial)
    \item This paper formally describes evaluation metrics used and explains the motivation for choosing these metrics. (partial)
    \item This paper states the number of algorithm runs used to compute each reported result. (no)
    \item Analysis of experiments goes beyond single-dimensional summaries of performance (e.g., average; median) to include measures of variation, confidence, or other distributional information. (yes)
    \item The significance of any improvement or decrease in performance is judged using appropriate statistical tests (e.g., Wilcoxon signed-rank). (no)
    \item This paper lists all final (hyper-)parameters used for each model/algorithm in the paper’s experiments. (partial)
    \item This paper states the number and range of values tried per (hyper-) parameter during development of the paper, along with the criterion used for selecting the final parameter setting. (partial)
\end{itemize}
\end{document}